\documentclass[pmlr]{jmlr}% new name PMLR (Proceedings of Machine Learning)

\RequirePackage{graphicx}
 \usepackage{booktabs}
\usepackage{subcaption}
\usepackage{csquotes}
\usepackage{float}
\usepackage{longtable}% for long tables
\makeatletter
\def\set@curr@file#1{\def\@curr@file{#1}} %temp workaround for 2019 latex release
\makeatother
\usepackage[load-configurations=version-1]{siunitx} % newer version
\usepackage{color} % for the command \textcolor
\usepackage{soul} % for the command \hl
\renewcommand{\cite}{\citep}

\theorembodyfont{\upshape}
\theoremheaderfont{\scshape}
\theorempostheader{:}
\theoremsep{\newline}

\jmlrvolume{298} %[VOLUME \# TBD]
\jmlryear{2025}
\jmlrworkshop{Machine Learning for Healthcare}

\title[ProtoECGNet]{ProtoECGNet: Case-Based Interpretable Deep Learning for Multi-Label ECG Classification with Contrastive Learning}

\author{ \\
\Name{Sahil Sethi}\textsuperscript{1,2} \Email{sethis@uchicago.edu} \\
\Name{David Chen}\textsuperscript{2} \Email{chend5@uchicago.edu} \\
\Name{Thomas Statchen}\textsuperscript{1,2} \Email{tstatchen@uchicago.edu} \\
\Name{Michael C. Burkhart}\textsuperscript{2} \Email{burkh4rt@uchicago.edu} \\
\Name{Nipun Bhandari}\textsuperscript{3} \Email{nibhandari@ucdavis.edu} \\
\Name{Bashar Ramadan}\textsuperscript{4,*} \Email{bashar.ramadan@bsd.uchicago.edu} \\
\Name{Brett Beaulieu-Jones}\textsuperscript{2,*} \Email{beaulieujones@uchicago.edu} \\
}

\begin{document}

\maketitle
\vspace{-3.5em} 
{\small
\noindent\textsuperscript{1}\textit{Pritzker School of Medicine, University of Chicago, IL, USA} \\
\noindent\textsuperscript{2}\textit{Center for Computational Medicine \& Clinical AI, Section of Biomedical Data Science, Department of Medicine, University of Chicago, IL, USA} \\
\noindent\textsuperscript{3}\textit{Division of Cardiovascular Medicine, Department of Internal Medicine, University of California Davis, CA, USA} \\
\noindent\textsuperscript{4}\textit{Section of Hospital Medicine, Department of Medicine, University of Chicago, IL, USA} \\
\\ \textit{*These authors contributed equally}
}

% \end{abstract}
\begin{abstract}
    Deep learning-based electrocardiogram (ECG) classification has shown impressive performance but clinical adoption has been slowed by the lack of transparent and faithful explanations. Post hoc methods such as saliency maps may fail to reflect a model’s true decision process. Prototype-based reasoning offers a more transparent alternative by grounding decisions in similarity to learned representations of real ECG segments—enabling faithful, case-based explanations. We introduce ProtoECGNet, a prototype-based deep learning model for interpretable, multi-label ECG classification. ProtoECGNet employs a structured, multi-branch architecture that reflects clinical interpretation workflows: it integrates a 1D CNN with global prototypes for rhythm classification, a 2D CNN with time-localized prototypes for morphology-based reasoning, and a 2D CNN with global prototypes for diffuse abnormalities. Each branch is trained with a prototype loss designed for multi-label learning, combining clustering, separation, diversity, and a novel contrastive loss that encourages appropriate separation between prototypes of unrelated classes while allowing clustering for frequently co-occurring diagnoses. We evaluate ProtoECGNet on all 71 labels from the PTB-XL dataset, demonstrating competitive performance relative to state-of-the-art black-box models while providing structured, case-based explanations. To assess prototype quality, we conduct a structured clinician review of the final model’s projected prototypes, finding that they are rated as representative and clear. ProtoECGNet shows that prototype learning can be effectively scaled to complex, multi-label time-series classification, offering a practical path toward transparent and trustworthy deep learning models for clinical decision support.
\end{abstract}

\section{Introduction}
\begin{figure}[t]
   \begin{center}
   \begin{tabular}{c} %% tabular useful for creating an array of images 
   \includegraphics [width=\textwidth]%[height=7.5cm]
   {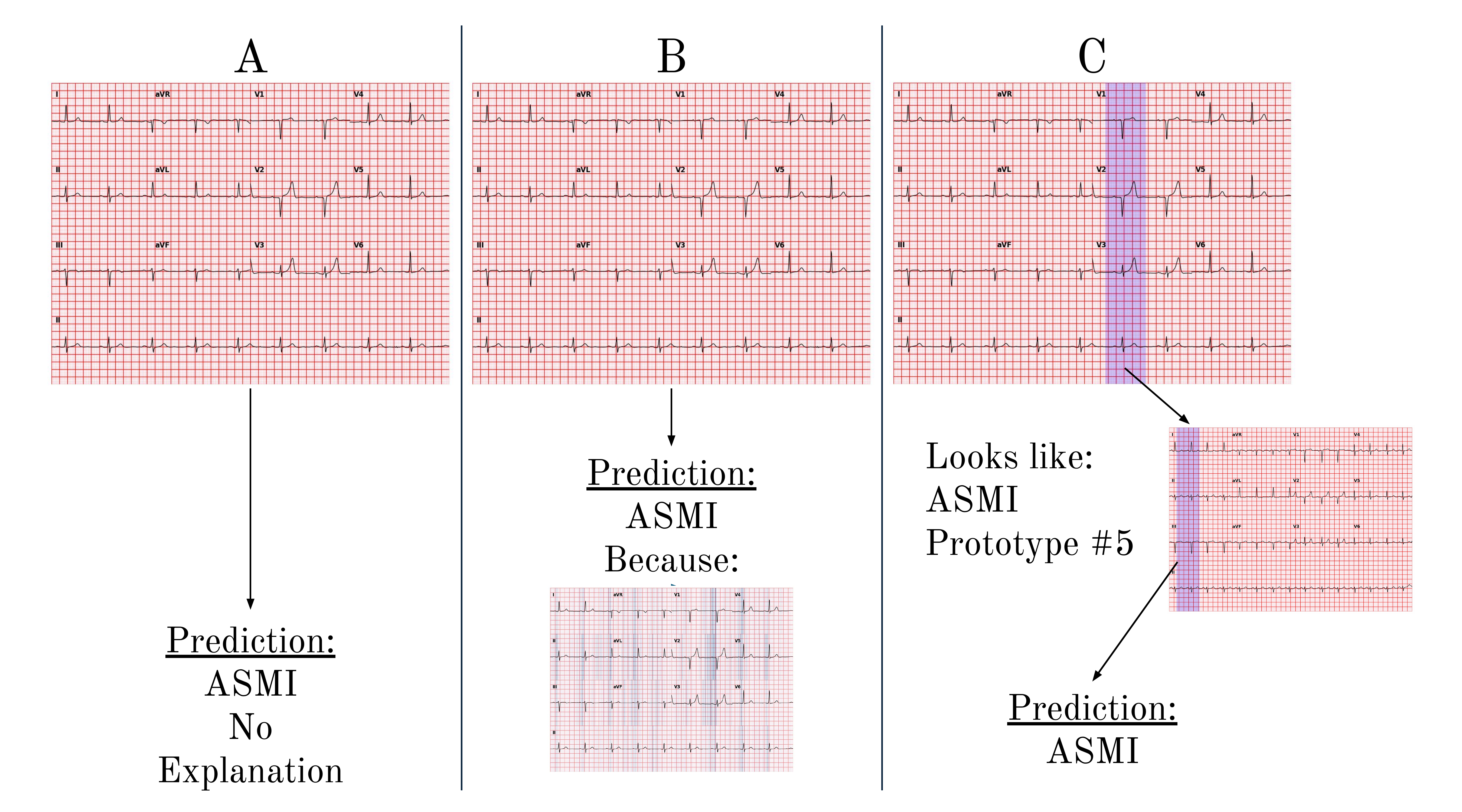}
   \end{tabular}
   \end{center}
   \caption[example] 
%>>>> use \label inside caption to get Fig. number with \autoref{}
   { \label{fig:introfig} 
Illustration of interpretability approaches for ECG classification. \textbf{(A)} Black-box models such as convolutional neural networks (CNNs) can achieve strong performance on diagnostic tasks, but provide no inherent explanation for their predictions. \textbf{(B)} Post hoc explainability methods, such as saliency maps, attempt to highlight input regions deemed important by the model after a prediction is made. However, these visualizations are not part of the model’s decision process and often fail to provide a meaningful explanation—simply indicating \enquote{where} the model looked does not explain \enquote{why} it made a decision. \textbf{(C)} Prototype-based models offer a self-explaining alternative: predictions are made by comparing a test input to a set of learned prototype vectors, each anchored to a real ECG segment. This enables case-based explanations that reflect the model’s actual classification logic. ASMI = anteroseptal myocardial infarction.

% Illustration of interpretability approaches for ECG classification. \textbf{(A)} Black-box models such as convolutional neural networks (CNNs) provide no inherent explanation for their predictions. \textbf{(B)} Post hoc explainability methods, such as saliency maps, attempt to highlight input regions deemed important by the model after a prediction is made. \textbf{(C)} Prototype-based models make predictions by comparing a test input to a set of learned prototype vectors, each anchored to a real ECG segment. This enables case-based explanations that reflect the model’s actual classification logic. ASMI = anteroseptal myocardial infarction.

}
\end{figure}

Deep learning (DL) has achieved strong performance across a wide range of diagnostic and predictive tasks in medicine \cite{aggarwal2021diagnostic, petmezas2022state, khera2024transforming, oliveira2023machine, barnes2023machine, sethi_toward_2025, sun_lesion-aware_2021}. In cardiology, DL-based electrocardiogram (ECG) interpretation is of great interest because ECGs are central to diagnosing many diseases like arrhythmias, myocardial infarction, and structural heart disease \cite{carrington_monitoring_2022, birnbaum_role_2014}. DL models have demonstrated strong, and in some cases, cardiologist-level performance in ECG classification \cite{hannun_cardiologist-level_2019, elias_deep_2022, he2023blinded, ouyang2024electrocardiographic, trivedi2025deep, yuan2023deep}, but their clinical deployment would be accelerated with increased transparency and trustworthiness in model predictions \cite{goettling_xecgarch_2024}. An example of a blackbox prediction for an ECG with an anteroseptal myocardial infarction (ASMI) is shown in \autoref{fig:introfig}A. 

Post hoc explainability methods—such as saliency maps and attention-based visualizations—are commonly used to interpret black-box deep learning models \cite{rudin_stop_2019, adebayo_sanity_2020}. However, these methods generate outputs that are not necessarily aligned with the model’s decision process. Prior studies have shown that saliency maps can be unstable, non-reproducible, and misaligned with human reasoning, particularly in medical domains \cite{adebayo_sanity_2020, alvarez-melis_towards_2018, turbe_evaluation_2023}. Most importantly, simply highlighting the parts of an ECG that the model focused on is not the same as explaining why it made a specific diagnosis—as illustrated in \autoref{fig:introfig}B. These limitations have led to growing interest in self-explaining models, where interpretability is embedded directly into the model architecture \cite{rudin_stop_2019, tonekaboni_what_2019}.

Prototype-based models classify inputs by comparing them to a small set of learned, class-associated vectors called \emph{prototypes}, each of which represents a localized region in the model’s latent space. During inference, predictions are based on the similarity between an input and these prototypes—effectively grounding decisions in similarity to learned representative examples from the training set. This allows the model to produce case-based explanations that are inherently faithful to its internal decision process (see \autoref{fig:introfig}C). Unlike saliency methods, prototype networks explicitly learn and are forced to use interpretable exemplars as part of their classification pipeline.  However, existing prototype-based models for ECG interpretation have been limited in scope. Prior work has focused primarily on single-label rhythm classification tasks and used only single- or dual-lead signals \cite{ming_interpretable_2019, xie_prototype_2024}. These approaches do not address the complexity of large-scale, multi-label ECG classification, where numerous cardiac abnormalities frequently co-occur and require diverse forms of temporal and spatial reasoning. 

In this work, we introduce \textbf{ProtoECGNet}, a prototype-based deep learning architecture for interpretable, multi-label ECG classification. ProtoECGNet is designed to mirror the reasoning processes used by clinicians during ECG interpretation, combining multiple prototype types aligned with temporal and spatial diagnostic patterns. Our approach enables structured, case-based explanations without sacrificing predictive performance, even in large-scale, multi-label classification tasks. Our key contributions include:

\begin{enumerate}
    \item \textbf{A customized prototype loss for multi-label ECG classification.} We build upon prior prototype learning objectives \cite{barnett_improving_2024, wang_interpretable_2021}, modifying the loss formulation to better accommodate multi-label supervision. Specifically, we retain and adapt the clustering and separation terms to account for label co-occurrence, and introduce a novel \emph{contrastive loss} designed for the multi-label prototype setting. Our contrastive loss encourages separation between prototypes assigned to rarely co-occurring diagnoses, while allowing prototypes for frequently co-occurring conditions to remain close in the latent space. This formulation structures the prototype geometry to reflect empirical co-occurrence patterns observed in the training data (e.g., Q-waves are a common co-occurrence across different forms of myocardial infarction) and improves classification performance.

    \item \textbf{A multi-branch model architecture aligned with clinical reasoning.} ProtoECGNet consists of three specialized prototype branches: (1) a \emph{1D rhythm model} with global prototypes designed to capture long-range temporal patterns, (2) a \emph{2D morphology model} with time-localized prototypes that leverage inter-lead spatial structure to identify focal waveform abnormalities, and (3) a \emph{2D global model} with full-duration, full-lead prototypes for detecting diffuse ECG abnormalities. Each branch was trained independently on a disjoint subset of diagnostic labels, and their outputs were either aggregated via macro-averaging or fused using a learned classifier trained to predict all 71 labels in the PTB-XL dataset.

    \item \textbf{Empirical evaluation of prototype quality by clinicians.} To assess the interpretability of the learned prototypes, we conducted a structured review in which two physicians independently rated all projected prototypes from the final model on multiple quality criteria, including clarity and class representativeness. This evaluation provides initial evidence that the model's explanations align with clinical expectations.
    
\end{enumerate}

% ProtoECGNet demonstrates that prototype-based reasoning can be effectively scaled to large-scale, multi-label ECG classification. By embedding case-based interpretability directly into both the model architecture and training objective, ProtoECGNet offers a practical path toward transparent and trustworthy deep learning systems for medical time-series analysis.

\subsection*{Generalizable Insights about Machine Learning in the Context of Healthcare}

We show that prototype-based interpretability depends on how prototypes are defined, trained, and aligned with clinical reasoning. By organizing diagnostic labels into rhythm, morphology, and global categories, and assigning each to a dedicated prototype branch, ProtoECGNet reflects the structure of expert ECG interpretation and enables more domain-tailored explanations. Further, we demonstrate that prototype learning can be extended to multi-label tasks through careful loss design: our contrastive formulation allows the model to preserve meaningful overlap between co-occurring diagnoses while maintaining discriminative structure in the latent space. Unlike prior prototype models limited to narrow single-label classification tasks, our approach supports full-spectrum ECG interpretation across 71 diagnostic labels, producing real ECG segment–based explanations without relying on post hoc methods. This shows that interpretable models can scale to realistic clinical tasks when their architecture and objectives are designed with domain constraints in mind.

\section{Related Work}
\subsection{Prototype-Based Learning}
\label{subsection:prototype_learning}
Prototype-based learning has emerged as a promising framework for interpretable machine learning, particularly following the introduction of ProtoPNet \cite{chen_this_2019}. Subsequent work has extended this paradigm in various directions: ProtoTree \cite{hase_interpretable_2019} structured prototypes hierarchically; TesNet \cite{wang_interpretable_2021} mapped prototypes to a hyperspherical latent space; ProtoPool \cite{rymarczyk_interpretable_2022} enabled soft sharing of prototypes across classes; and ProtoConcepts \cite{ma_this_2023} combined prototypes with concept-based reasoning. These models have demonstrated interpretability benefits in imaging tasks, including breast cancer classification \cite{barnett_case-based_2021}, brain tumor detection \cite{wei_mprotonet_2023}, and chest X-ray analysis \cite{kim_xprotonet_2021}. However, most of this literature assumes mutually exclusive class labels and focuses on static 2-D images, limiting direct applicability to sequential medical data.

Prototype learning for time-series data remains underexplored. One adaptation to EEG classification uses a 2D CNN-based prototype model, treating multichannel signals as images and validating interpretability through a structured clinician study \cite{barnett_improving_2024}. However, this approach used only global prototypes and addressed a single-label, multi-class classification task. Existing prototype-based ECG models, such as ProSeNet \cite{ming_interpretable_2019} and PahNet \cite{xie_prototype_2024}, focus narrowly on rhythm detection using single- or dual-lead inputs, and neither supports multi-label classification. Moreover, PahNet does not include a prototype projection mechanism, preventing alignment between learned prototypes and real ECG segments—limiting its explanatory usefulness in clinical settings.

The most closely related work is xECGArch \cite{goettling_xecgarch_2024}, which introduces a dual-CNN model to separately capture short-term (morphological) and long-term (rhythmic) patterns in ECGs. However, xECGArch does not project prototypes onto real ECG segments and instead relies on post hoc saliency maps for interpretation, limiting its ability to provide case-based explanations. It is also restricted to binary atrial fibrillation detection using single-lead inputs. In contrast, ProtoECGNet is explicitly designed for multi-label, 12-lead ECG interpretation. It includes three clinically inspired prototype branches tailored to rhythm, morphology, and global abnormalities, with each prototype anchored to a real training segment for faithful explanation. The latent space is further structured using a custom contrastive loss that reflects real-world label co-occurrence. Our work extends prototype learning to a more complex and clinically realistic setting.

\subsection{Contrastive Learning}
\label{subsection:contrastive}

While prototype learning offers an inherently interpretable model structure, it does not ensure that the learned prototypes reflect meaningful diagnostic variation—particularly in multi-label settings, where some conditions routinely co-occur. To address this, we introduce a contrastive loss tailored to multi-label prototype learning, designed to shape the prototype similarity space in alignment with label relationships observed in the training data.

Our approach is inspired by supervised contrastive learning, which has been widely used to structure encoder representations. SupCon \cite{khosla_supervised_2021} and SimCLR \cite{chen_simple_2020} promote similarity among positive pairs and dissimilarity among negatives using log-softmax objectives over large batches. Multi-label extensions, such as MulSupCon \cite{zhang_multi-label_2024}, weight similarity based on label overlap. However, these instance-level methods are not applicable to our setting, where the goal is to organize a fixed set of prototype vectors with known class assignments—not to embed instances.

More relevant is the Joint Supervised Contrastive Loss (JSCL) proposed by~\citet{lin_effective_2023}, which introduces a contrastive loss for multi-label classification by encouraging proximity between latent representations of examples with shared labels and pushing apart those with disjoint labels. JSCL operates on input embeddings and uses a log-ratio formulation. We adapt this idea for the prototype learning setting in two key ways. First, we apply the loss directly to the prototype vectors in the latent space, leveraging their fixed prototype-to-class assignments to define positive and negative pairs. Second, we replace the log-ratio formulation with a mean-difference objective: the average similarity between prototype pairs assigned to overlapping labels is contrasted against the average similarity of those with disjoint label sets. This formulation is well-suited to our structured prototype setting and encourages the latent space to reflect realistic co-occurrence patterns.

\section{Methods}
\subsection{Dataset and Label Grouping}
We used the PTB-XL dataset, a publicly available collection of 21,799 10-second, 12-lead ECGs from 18,869 patients sampled at both 500 Hz and 100 Hz \cite{wagner_ptb-xl_2020}. Each ECG was annotated with one or more of 71 SCP-ECG labels spanning arrhythmias, conduction disorders, infarction patterns, and morphological abnormalities. We retained the original dataset split from \citet{wagner_ptb-xl_2020}—using folds 1-8 for training and fold 9 for validation. We reported performance metrics on fold 10, using it as a hold-out test set. To align with the dataset's benchmarks, macro-AUROC from \citet{strodthoff_deep_2021} was chosen as the primary performance metric. Appendix \ref{app:ecg_visualization} details how ECGs were visualized for figures and prototype review. 

The dataset contains several label groupings (e.g., diagnostic superclasses), but these were unsuitable for our desired training process. Consequently, two physicians—one board-certified in cardiology and the other in internal medicine—grouped the 71 labels into three clinically meaningful prototype categories based on the type of visual reasoning required for diagnosis: \textbf{(1) rhythm-based diagnoses}—require temporal pattern analysis across full-length ECG signals, often discernible from a single lead; \textbf{(2) morphology-based diagnoses}—require localized waveform shape or inter-lead comparisons over short time intervals; \textbf{(3) global diagnoses}—require full-lead patterns spanning the full ECG duration. In total, 16 diagnoses were grouped into the 1D rhythm branch, 52 into the 2D morphology branch, and 3 into the 2D global branch. These groupings are detailed in Appendix \ref{app:diag_groups}. 

\subsection{Preprocessing}
As the lower end of the frequency range for a normal ECG is 0.5 Hz \cite{zheng_12-lead_2020}, we applied a first-order Butterworth high-pass filter with a 0.5~Hz cutoff. No low-pass filter was applied as we used the 100 Hz samples from the dataset. 

\subsection{Model Architecture}
\label{sec:architecture}
\begin{figure}
\centering
\includegraphics[width=\textwidth]{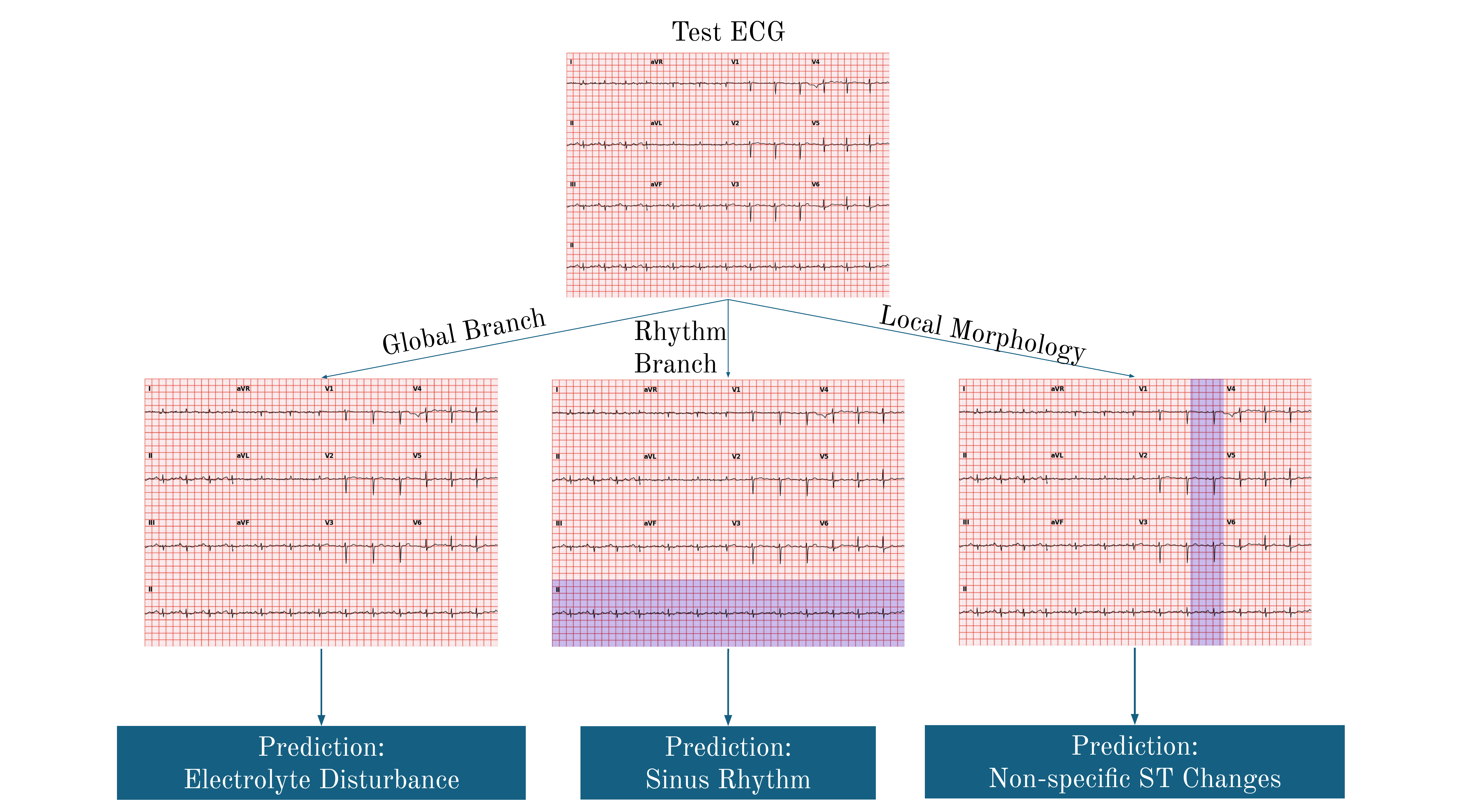}
\caption{\label{fig:methods} Multi-branch approach. See \autoref{fig:app_methods} for detailed architectural information.}
\end{figure}

ProtoECGNet consisted of three independent prototype-based branches, each specialized for a distinct type of ECG diagnostic reasoning: rhythm-based, morphology-based, and global abnormalities (see \autoref{fig:methods} and \autoref{fig:app_methods}). Each branch processed the same raw 12-lead ECG input of shape $(12 \times 1000)$ but applied a different label subset.

\paragraph{1D Rhythm Branch.}
This branch was designed to capture long-range temporal dependencies characteristic of rhythm abnormalities. We adopted the ResNet1D-18 architecture from \citet{strodthoff_deep_2021}, operating on input ECGs of shape $(12 \times 1000)$. The architecture consisted of an initial strided convolution, followed by four residual blocks with increasing channel depth. Feature maps were pooled with an adaptive average pooling layer to produce fixed-length latent representations across leads. These representations formed the prototype matching space for the 1D rhythm model.

\paragraph{2D Morphology Branch.} 
To detect localized waveform abnormalities that depend on spatial relationships across ECG leads—such as ST elevation or pathological Q waves—we used a 2D convolutional neural network that treated the 12-lead ECG as a spatial-temporal matrix. Input ECGs were shaped as $(1 \times 12 \times 1000)$, where the vertical dimension corresponded to the 12 leads and the horizontal dimension to time. The model used a modified ResNet18 \cite{he_deep_2015} backbone in which the first convolutional layer was adapted to accept single-channel 2D inputs with a filter size of $(12 \times 7)$. Pretrained ImageNet weights were loaded for the remaining layers. To preserve temporal resolution, the global average pooling layer was removed, resulting in a final feature map of shape $(512 \times 1 \times 32)$, where the temporal axis was downsampled from 1000 to 32 steps. This branch used \textit{partial prototypes} that spanned a small temporal window (3 units, approximately 0.94 seconds at 100~Hz) and were applied in a sliding fashion across the latent time dimension. During inference, the model computed similarity scores between each partial prototype and all possible time-localized windows of the test ECG’s latent feature map. Top-$k$ pooling (with $k=5$) was then applied to aggregate the $k$ most activated positions into a final similarity score per prototype. 

\paragraph{2D Global Branch.} 
Diagnoses that required global ECG interpretation—such as electrolyte disturbances—were handled using the same 2D CNN architecture as the 2D morphology branch. Inputs had shape $(1 \times 12 \times 1000)$ and yielded latent maps of shape $(512 \times 1 \times 32)$. However, this branch used \textit{global prototypes} spanning all leads and the full time axis. 

\paragraph{Prototype Layer.}
All prototype matching occurred in the latent feature space of the CNN feature extractor backbones. Each branch contained its own prototype layer with $P$ learnable prototypes assigned to specific class labels. Let $z_i \in \mathbb{R}^D$ be a patch from the latent feature map and $p_j \in \mathbb{R}^D$ a prototype. We defined the similarity between them as:
\begin{equation}
\label{eq:similarity}
\text{S}(z_i, p_j) = 
\left\langle 
\frac{a \cdot z_i}{\|z_i\|_2}, 
\frac{p_j}{\|p_j\|_2} 
\right\rangle
\end{equation} where $\langle \cdot, \cdot \rangle$ denotes the dot product and the scaling factor $a$ adjusts the magnitude of the similarity score based on the latent dimensionality. This cosine similarity variant, adapted from \citet{barnett_improving_2024}, was used to compute similarity between each prototype and regions of the latent feature map. For \textit{partial prototypes} (used in the morphology branch), each prototype slid across the temporal axis of the latent feature map and produced a similarity score at each time step. We then applied top-$k$ pooling (with $k=5$) across these scores to retain the $k$ highest activations, and computed their average to obtain a single similarity score per prototype per input. For \textit{global prototypes} (used in the rhythm and 2D global branches), each prototype spanned the entire temporal dimension of the latent space, yielding a single similarity score per input without the need for pooling.

\subsection{Prototype Loss Function}
\label{sec:loss}

Each branch was trained using a composite loss function:
\begin{equation}
\mathcal{L}_{\text{total}} =
\mathcal{L}_{\text{BCE}} + 
\lambda_{\text{clst}} \cdot \mathcal{L}_{\text{clst}} +
\lambda_{\text{sep}} \cdot \mathcal{L}_{\text{sep}} +
\lambda_{\text{div}} \cdot \mathcal{L}_{\text{div}} +
\lambda_{\text{cntrst}} \cdot \mathcal{L}_{\text{cntrst}}
\end{equation}
where each \( \lambda \) is a tunable hyperparameter controlling the strength of its loss component.

\paragraph{Binary Cross-Entropy Loss.}
\begin{equation}
\mathcal{L}_{\text{BCE}} =
- \frac{1}{N} \sum_{i=1}^{N} \sum_{j=1}^{C} 
w_j \left[ y_{ij} \log \sigma(z_{ij}) + (1 - y_{ij}) \log (1 - \sigma(z_{ij})) \right]
\end{equation}

This loss penalized incorrect predictions for each class independently in a multi-label setting. Here, \( y_{ij} \in \{0,1\} \) is the ground truth label for sample \( i \) and class \( j \); \( z_{ij} \) is the predicted logit; \( \sigma(\cdot) \) is the sigmoid function; and \( w_j \) is a class weight.

\paragraph{Clustering Loss.}
\begin{equation}
\mathcal{L}_{\text{clst}} =
- \frac{1}{N} \sum_{i=1}^{N} 
\max_{p \in \mathcal{P}_i^+} S_{ip}
\end{equation}
This term encouraged each prototype to have high similarity to at least one input that shared its assigned class label. \( S_{ip} \) is the similarity between input \( i \) and prototype \( p \), and \( \mathcal{P}_i^+ \) denotes the set of prototypes assigned to any of the labels present in sample \( i \). Note that this term differed from \citet{chen_this_2019} in that an exact label match was not required. 

\paragraph{Separation Loss.}
\begin{equation}
\mathcal{L}_{\text{sep}} =
\frac{1}{N} \sum_{i=1}^{N} 
\max_{p \in \mathcal{P}_i^-} S_{ip}
\end{equation}
This term penalized high similarity between a training sample and prototypes not associated with any of its labels. \( \mathcal{P}_i^- \) is the set of prototypes whose class assignments are entirely disjoint from the ground-truth labels of sample \( i \). 

\paragraph{Orthogonality Loss.}
\begin{equation}
\mathcal{L}_{\text{div}} = 
\left\| \mathbf{P} \mathbf{P}^\top - \mathbf{I} \right\|_F^2
\end{equation}
This discouraged redundancy among prototypes. Let \( \mathbf{P} \in \mathbb{R}^{P \times D} \) be the matrix of flattened, row-normalized prototype vectors, and \( \mathbf{I} \) the identity matrix. The Frobenius norm measures deviation from orthonormality. This term was directly adapted from \citet{barnett_improving_2024}.

\paragraph{Contrastive Loss.} 

Together, our modifications to the clustering and separation loss terms extended their logic from requiring an exact label match between a prototype and a training example to allowing prototypes to be attracted to any example that contains their assigned class—regardless of what other labels are present. This enabled clustering in the presence of multi-label supervision and implicitly supported overlap among frequently co-occurring diagnoses. However, these terms do not directly structure the relationships between prototypes themselves. To more explicitly shape the geometry of the prototype space based on diagnostic co-occurrence, we introduced a contrastive loss applied at the level of prototype-prototype similarity. This encouraged prototypes assigned to frequently co-occurring classes to remain similar in the latent space, while discouraging similarity between prototypes associated with rarely co-occurring diagnoses. 

\begin{equation}
\mathcal{L}_{\text{cntrst}} = 
- \frac{1}{\sqrt{P}} \left( 
\frac{\sum_{i,j} C_{ij} \cdot \text{S}(p_i, p_j)}{\sum_{i,j} C_{ij}} - 
\frac{\sum_{i,j} (1 - C_{ij}) \cdot \text{S}(p_i, p_j)}{\sum_{i,j} (1 - C_{ij})}
\right)
\end{equation}

Our contrastive loss was inspired by the Jaccard Similarity Contrastive Loss (JSCL) proposed by \citet{lin_effective_2023} in the context of multi-label text classification. JSCL weights the contrastive objective between sample pairs by the Jaccard similarity of their label sets, allowing for a soft notion of positive and negative pairs in multi-label settings. We adopted this core idea—using Jaccard similarity to scale pairwise contrastive terms—and adapted it to prototype learning by computing similarity between learned prototype vectors. Specifically, we precomputed a Jaccard-based co-occurrence matrix \( C \in [0,1]^{P \times P} \) over the training label set and used it to weight the pairwise similarity score between prototypes. \( S({p_i,p_j}) \) is the similarity between prototype $i$ and prototype $j$, and $P$ is the number of prototypes.

While our contrastive loss encouraged prototypes of frequently co-occurring classes to remain nearby in the latent space, the model still enforced class-specificity in its decision-making in several ways. First, the classification loss was computed independently for each class using sigmoid activation and binary cross-entropy, ensuring that incorrect predictions were directly penalized regardless of prototype proximity. Second, we included an orthogonality regularization term to promote diversity among prototypes and reduce redundancy. Third, both the branch-specific classifiers and the fusion classifier underwent an L1-constrained convex optimization step after prototype projection, which minimized unnecessary weights and suppressed prototypes that did not meaningfully contribute to final predictions. Together, these mechanisms ensured that similarity in the latent space did not translate into redundant or incorrect activations in the prediction pipeline.

\subsection{Training Procedure}

Prototype-based models underwent a three-stage training procedure (see Appendix \ref{sec:training}). Stage 1 (joint training) and Stage 2 (prototype projection) were repeated as needed (until validation AUC no longer increased) to allow the prototypes to better converge on class-representative features. Stage 3 (fusion classifier training) was conducted only once, after freezing the prototype branches. 

\subsection{Manual Evaluation of Prototype Quality}
\label{sec:userstudy}
To assess the interpretability and perceived clinical utility of the learned prototypes, we conducted a structured review with two practicing physicians. The evaluation focused on the final ProtoECGNet model trained with the full prototype loss suite, including contrastive regularization. Details on the graphical user interface (GUI) used for clinician review and a screenshot of the interface are provided in Appendix \ref{app:interface}.

The reviewers included a board-certified internist and a board-certified cardiologist. This pairing was selected to reflect both generalist and specialist clinical perspectives. Each reviewer was asked to independently evaluate all available prototypes from the final contrastive model using a lightweight web-based interface. Prototype quality was assessed along two criteria: 1.) Representativeness. Does the prototype reflect a typical or defining presentation of the assigned diagnostic class? 2.) Clarity. Is the ECG signal in the prototype clean and interpretable, or is it obscured by noise or artifacts that make interpretation difficult?

Each criterion was rated on a 1–5 Likert scale. Reviewers were instructed to score each prototype independently based on the projected ECG segment (see Appendices \ref{app:interface} and \ref{app:ecg_visualization} for additional details). No test cases or model predictions were presented. ECGs that were identified as having label errors by the clinicians were excluded. We report the mean scores and 95\% confidence intervals across reviewers for each evaluation criterion.

\section{Results \& Discussion}
\label{sec:results_discussion}

We evaluate ProtoECGNet on classification performance and prototype interpretability. All results are reported on the held-out PTB-XL test set (fold 10). Macro-AUROC is used as the primary evaluation metric throughout to allow comparison to the PTB-XL benchmarking study \cite{strodthoff_deep_2021}. However, we also report weighted AUROCs to allow for reporting 95\% confidence intervals (CIs) (bootstrapped with 10,000 samples). Weighted AUROCs were required because bootstrapping creates numerous resamplings of the test set, and some of the resamplings do not contain positive examples for all classes (a handful of classes have just one positive example in the test set; sometimes, that one positive example is not included in the resampling). We calculate AUROC for each class in each resampling. For the resamplings that do not contain a positive example of a given class, the AUROC for that class is not defined. The macro-AUROC forms an unweighted average over the AUROCs for each class, and so is also undefined for the resamplings that are missing positive examples of one of the classes. The weighted AUROC averages over the AUROCs in proportion to the number of positive examples in the resampling, and so can safely ignore the AUROCs that are undefined. Class-specific AUROCs for our final model and counts of positive examples for each class are provided in Appendix \ref{app:class_specific_aucs}.

\subsection{Does explicitly modeling rhythm, morphology, and global abnormalities improve performance over using a single prototype type?}

\textbf{Experiment:} To evaluate whether aligning prototype types with clinical reasoning modalities improves performance, we compare models trained on all 71 PTB-XL labels using (1) a single prototype type (e.g., only 1D or 2D global), and (2) the full ProtoECGNet multi-branch model. Prototype-based models are trained with and without contrastive loss, and the multi-branch outputs are fused using either macro-averaging or a learned fusion classifier.

\textbf{Results:} The best-performing single-branch model was the 2D partial prototype model with contrastive loss, achieving a macro-AUROC of 0.9091 (see \autoref{tab:macro_auc_results}). The full ProtoECGNet model—combining 1D, 2D partial, and 2D global branches—achieved the highest overall performance when using a learned fusion classifier, with a macro-AUROC of 0.9132. This exceeds all single-branch models and approaches the performance of the best models in the PTB-XL benchmarking study (macro-AUROC 0.925 for single model; 0.929 for ensemble) \cite{strodthoff_deep_2021}. The same trends were observed when using weighted AUROC (see \autoref{tab:weighted_auc_results}); the fusion classifier outperformed single-branch models and achieved 0.9066 (95\% CI: 0.9000-0.9128).

\begin{table}[h]
\centering
\caption{\label{tab:macro_auc_results} Macro-AUROC across branch-specific, single-branch, and multi-branch settings for ProtoECGNet.}
\resizebox{\linewidth}{!}{
\begin{tabular}{lllccc}
\toprule
\textbf{Setting} & \textbf{Model (Label Set)} & \textbf{Black-box} & \textbf{No Contrastive} & \textbf{w/ Contrastive} \\
\midrule
{Branch-Specific Labels} 
& Rhythm Branch (16 labels) & 0.9403 & 0.8903 & 0.9064 \\
& Morphology Branch (52 labels) & 0.8872 & 0.8533 & 0.9051 \\
& Global Branch (3 labels) & 0.8649 & 0.8362 & 0.8667 \\
\midrule
{Full 71-Label, Single Branch}
& 1D Prototype Model & \textbf{0.9250} & 0.8646 & 0.8977 \\
& 2D Partial Prototype Model & N/A & 0.8873 & 0.9091 \\
& 2D Global Prototype Model & 0.8990 & 0.8681 & 0.9074 \\
\midrule
{Full 71-Label, Multi-Branch}
& Macro Aggregation & 0.8982 & 0.8609 & 0.9038 \\
& Fusion Classifier & N/A & 0.8855 & \textbf{0.9132} \\
\bottomrule
\end{tabular}
}
\end{table}

\begin{table}[h]
\centering
\caption{\label{tab:weighted_auc_results} Weighted AUROC with bootstrapped 95\% confidence intervals across branch-specific, single-branch, and multi-branch settings.}
\resizebox{\linewidth}{!}{
\begin{tabular}{lllccc}
\toprule
\textbf{Setting} & \textbf{Model (Label Set)} & & \textbf{Black-box} & \textbf{No Contrastive} & \textbf{w/ Contrastive} \\
\midrule
{Branch-Specific} 
& Rhythm Branch (16 labels) & & 0.8919 (0.8757, 0.9071) & 0.8762 (0.8616, 0.8901) & 0.8853 (0.8708, 0.8991) \\
& Morphology Branch (52 labels) & & 0.8996 (0.8930, 0.9057) & 0.8791 (0.8727, 0.8855) & 0.8996 (0.8931, 0.9059) \\
& Global Branch (3 labels) & & 0.8307 (0.8137, 0.8470) & 0.6981 (0.6767, 0.7197) & 0.9039 (0.8906, 0.9164) \\
\midrule
{Single-Branch}
& 1D Prototype Model & & \textbf{0.9081 (0.9012, 0.9147)} & 0.8108 (0.8025, 0.8188) & 0.8857 (0.8782, 0.8930) \\
& 2D Partial Prototype Model & & N/A & 0.8605 (0.8526, 0.8684) & 0.8743 (0.8666, 0.8819) \\
& 2D Global Prototype Model & & 0.8932 (0.8859, 0.9002) & 0.8589 (0.8505, 0.8669) & 0.8916 (0.8841, 0.8992) \\
\midrule
{Multi-Branch}
& Macro Aggregation & & 0.8855 (0.8779, 0.8926) & 0.8486 (0.8413, 0.8557) & 0.8950 (0.8882, 0.9016) \\
& Fusion Classifier & & N/A & 0.8800 (0.8728, 0.8872) & \textbf{0.9066 (0.9000, 0.9128)} \\
\bottomrule
\end{tabular}
}
\end{table}

\textbf{Discussion:} These results support the hypothesis that modeling distinct diagnostic reasoning types using specialized prototype branches can improve performance. While the contrastive 2D partial prototype model (0.9091 macro-AUROC) was the strongest among single-branch models, the multi-branch ProtoECGNet fusion classifier achieved the highest macro-AUROC (0.9132) and weighted AUROC (0.9066). This demonstrates that structuring the architecture to reflect clinical reasoning modalities does not compromise—and may modestly improve—overall diagnostic performance.

\subsection{Does contrastive prototype loss improve diagnostic performance?}

\textbf{Experiment:} We evaluate the effect of our proposed contrastive prototype loss across three settings: (1) branch-specific models trained on disjoint label subsets (rhythm, morphology, global), (2) single-branch models trained on all 71 labels, and (3) the full multi-branch model. Each prototype-based model is trained with and without contrastive loss and compared to its corresponding baseline.

\textbf{Results:} Contrastive loss improved macro-AUROC in all settings. In branch-specific models, the 2D global prototype model improved from 0.8362 to 0.8667, the rhythm model from 0.8903 to 0.9064, and the morphology model from 0.8533 to 0.9051. In the full-label (single-branch) setting, contrastive loss improved the 1D prototype model from 0.8646 to 0.8977, the 2D partial prototype model from 0.8873 to 0.9091, and the 2D global prototype model from 0.8681 to 0.9074. For the multi-branch model, contrastive loss improved macro-AUROC from 0.8855 to 0.9132.

Weighted AUROCs showed similar trends. In the multi-branch fusion model, contrastive loss improved performance from 0.8800 (95\% CI: 0.8728-0.8872) to 0.9066 (95\% CI: 0.9000-0.9128), with no overlap in confidence intervals. The 2D global branch-specific model also showed a substantial improvement, from 0.6981 (95\% CI: 0.6767-0.7197) to 0.9039 (95\% CI: 0.8906-0.9164), again with non-overlapping intervals. The morphology branch-specific model increased from 0.8791 (95\% CI: 0.8727-0.8855) to 0.8996 (95\% CI: 0.8931-0.9059), also without overlap. In contrast, the rhythm branch-specific model improved from 0.8762 (95\% CI: 0.8616-0.8901) to 0.8853 (95\% CI: 0.8708-0.8991), but with overlapping intervals. The single-branch 1D prototype model improved from 0.8108 (95\% CI: 0.8025-0.8188) to 0.8857 (95\% CI: 0.8782-0.8930), and the 2D global prototype model from 0.8589 (95\% CI: 0.8505-0.8669), both without overlapping intervals. However, confidence intervals slightly overlapped for the single-branch 2D partial prototype model, which improved from 0.8605 (95\% CI: 0.8526-0.8684) to 0.8743 (95\% CI: 0.8666-0.8819).

\textbf{Discussion:} Contrastive prototype loss improved macro-AUROC across all model settings. Weighted AUROC was also higher for all model configurations, with most settings showing non-overlapping confidence intervals. Notably, contrastive regularization improved both the macro- and weighted AUROC for the final multi-branch ProtoECGNet—supporting the importance of class co-occurrence-informed regularization in multi-label prototype learning.

\subsection{How does ProtoECGNet compare to black-box baselines?}

\textbf{Experiment:} For each model variant—1D, 2D partial, 2D global, and multi-branch—we compare a black-box ResNet18 baseline with the corresponding prototype models (with and without contrastive loss). This allows us to assess whether our interpretable architecture sacrifices diagnostic performance.

\textbf{Results:} The black-box models achieved macro-AUROCs of 0.9403 (1D rhythm subset), 0.8872 (2D morphology subset), and 0.8649 (2D global subset) when trained on their respective label subsets. When combined, their macro-averaged performance was 0.8982. When trained directly on all 71 labels, the 1D black-box model achieved 0.9250 macro-AUROC and the 2D black-box model achieved 0.8990. Among prototype models trained on all 71 labels, the contrastive 2D partial prototype model achieved 0.9091, and the final contrastive multi-branch fusion model achieved 0.9132.

In terms of weighted AUROC, the 1D black-box model achieved 0.9081 (0.9012, 0.9147), the 2D black-box model 0.8932 (0.8859, 0.9002), and the contrastive fusion model 0.9066 (0.9000, 0.9128). The confidence intervals for the 1D black-box model and the final contrastive fusion model overlapped.

\textbf{Discussion:} The 1D black-box model achieved the highest overall performance (macro-AUROC 0.9250; weighted AUROC 0.9081), matching the strongest non-ensemble model reported in the PTB-XL benchmarking study \cite{strodthoff_deep_2021}. The contrastive multi-branch prototype model achieved slightly lower macro-AUROC (0.9132), but its weighted AUROC of 0.9066 had overlapping confidence intervals with the black-box model, indicating statistically comparable performance. These results demonstrate that interpretable, prototype-based models can retain diagnostic accuracy while providing transparent, case-based explanations.

\subsection{Are the learned prototypes clinically meaningful?}
\label{sec:prototype-eval}

\textbf{Experiment:} To assess interpretability, we conducted a structured review of all projected prototypes from the final contrastive model. Two physicians—one with board certification in cardiology and one in internal medicine—independently rated each prototype on a 1–5 Likert scale for two criteria: \textit{representativeness} (how well the prototype exemplifies the assigned class) and \textit{clarity} (how visually interpretable and artifact-free the segment is). These criteria were adapted from literature on concept-based interpretability \citep{alvarez-melis_towards_2018, ghorbani_towards_2019}. Specifically, representativeness reflects the \enquote{meaningfulness} and \enquote{importance} criteria from \citet{ghorbani_towards_2019}, which define whether a concept is both human-interpretable and relevant to the model’s prediction. Clarity incorporates aspects of \enquote{coherency} from \citet{ghorbani_towards_2019}, and \enquote{explicitness} from \citet{alvarez-melis_towards_2018}, capturing how immediately understandable and visually consistent an explanation is.

\textbf{Results:} Average scores from the clinicians for both criteria are shown in \autoref{tab:clinician-scores}. In brackets, 95\% confidence intervals (CIs) are presented.

\begin{table}[h]
\centering
\caption{Average prototype quality scores from structured clinician review (1–5 scale).}
\label{tab:clinician-scores}
\begin{tabular}{lcc}
\toprule
\textbf{Reviewer} & \textbf{Representativeness (95\% CI)} & \textbf{Clarity (95\% CI)} \\
\midrule
Cardiologist & 4.29 [4.22, 4.35] & 4.48 [4.42, 4.54] \\
Internist    & 3.59 [3.52, 3.66] & 4.73 [4.69, 4.77] \\
\bottomrule
\end{tabular}
\end{table}

\textbf{Discussion:} These results suggest ProtoECGNet learns meaningful ECG patterns which align with clinician expectations. High average scores for both representativeness and clarity indicate that the prototypes are visually interpretable and diagnostically appropriate. This review provides initial validation that the model generates human-interpretable prototypes while maintaining predictive performance across diverse diagnostic categories but it does not assess downstream utility. Future work plans to include this assessment as a larger user study. 

\subsection{How does ProtoECGNet support case-based explanations?}
\label{sec:case-examples}

\textbf{Experiment:} To evaluate the interpretability of ProtoECGNet’s predictions, we qualitatively examined projected prototypes for representative test examples. For each test case, we visualized the most strongly activated prototype and its corresponding training ECG. 

\begin{figure}[t]
   \begin{center}
   \begin{tabular}{c} %% tabular useful for creating an array of images 
   \includegraphics [width=\textwidth]%[height=7.5cm]
   {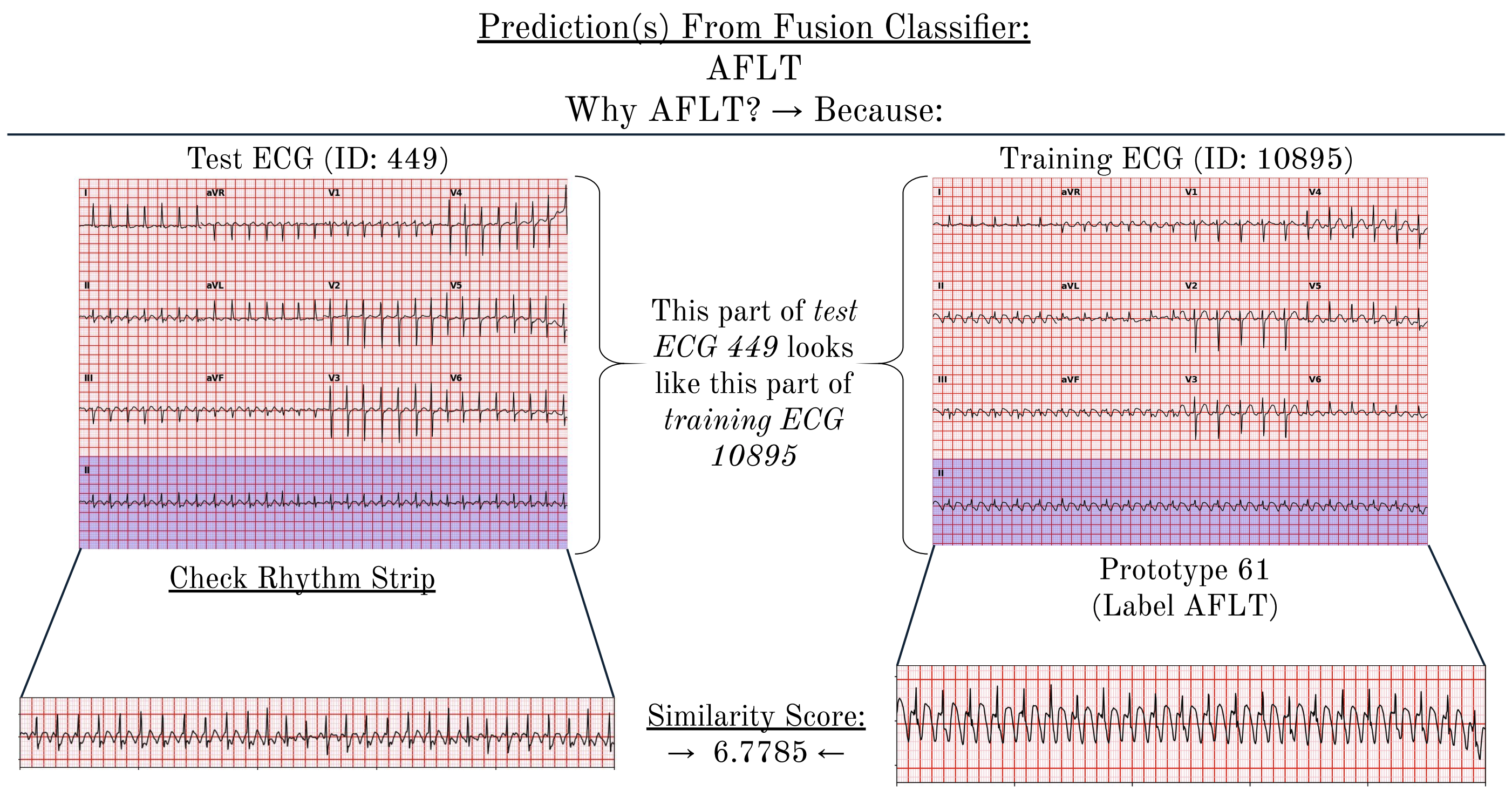}
   \end{tabular}
   \end{center}
   \caption[example] 
%>>>> use \label inside caption to get Fig. number with \autoref{}
   { \label{fig:cat1_case} Case-based explanation for atrial flutter (AFLT) predicted by the fusion classifier. The model predicts AFLT for test ECG 449 based on high similarity to prototype 61, which was projected onto training ECG 10895. The top row displays the full 12-lead ECGs for both examples, with rhythm strips (lead II) highlighted in blue to guide interpretation. The bottom row provides a zoomed-in view of these rhythm strips.}
\end{figure}

\begin{figure}[t]
   \begin{center}
   \begin{tabular}{c} %% tabular useful for creating an array of images 
   \includegraphics [width=\textwidth]%[height=7.5cm]
   {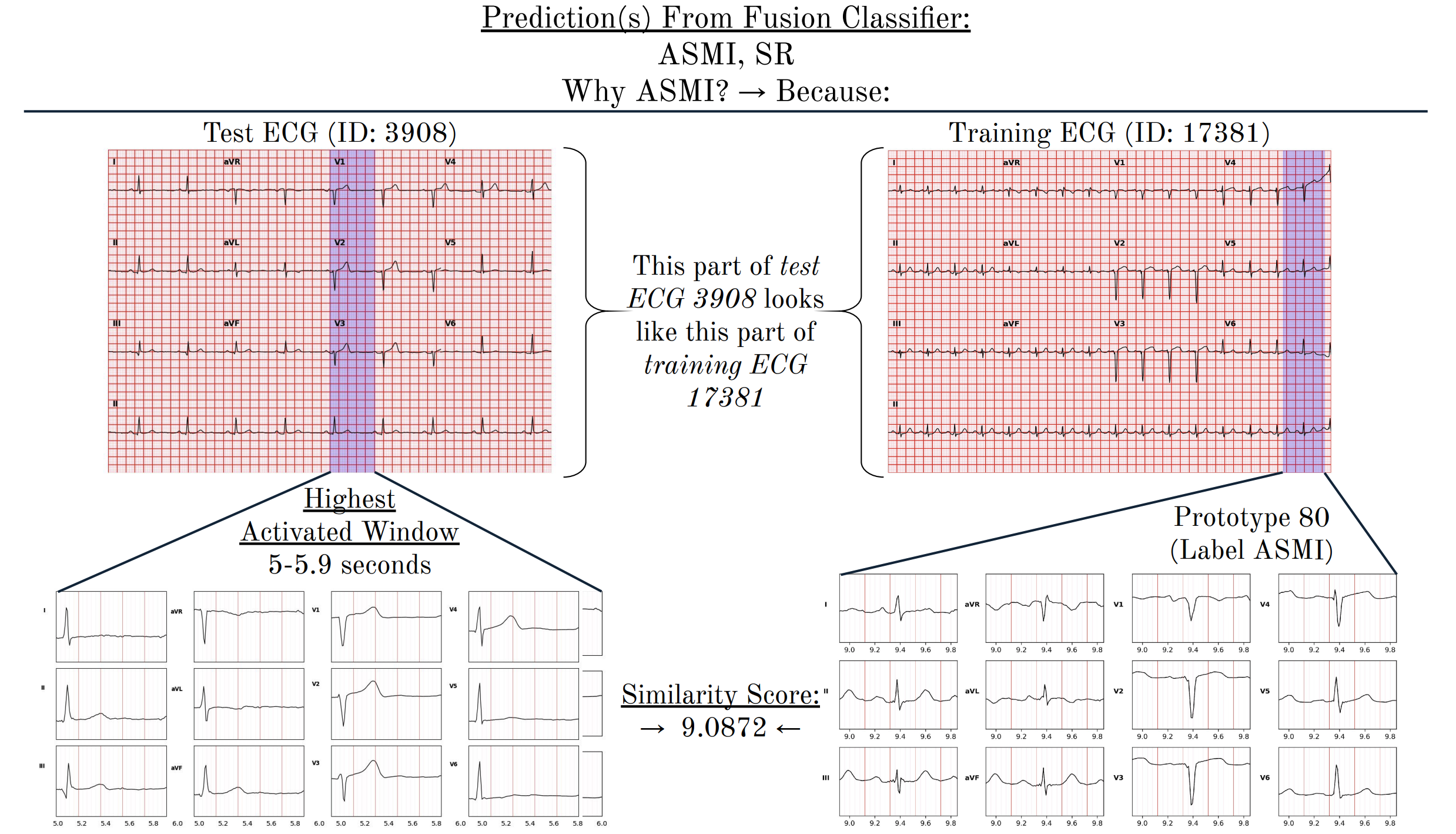}
   \end{tabular}
   \end{center}
   \caption[example] 
%>>>> use \label inside caption to get Fig. number with \autoref{}
   { \label{fig:cat3_case} Case-based explanation for anteroseptal myocardial infarction (ASMI) predicted by the fusion classifier. The model predicts ASMI for test ECG 3908, citing strong similarity to prototype 80, which was projected onto a latent patch from training ECG 17381. The top row shows the full 12-lead ECGs for both test and training examples, with the activated region highlighted in blue (5–5.9 seconds for the test ECG and 8.9–9.8 seconds for the prototype). The bottom row zooms into these regions to show all 12 leads. The model appears to have identified a match based on ST-segment elevations in anterior leads (e.g., V2–V4), with a high similarity score of 9.0872.}
\end{figure}

\begin{figure}[t]
   \begin{center}
   \begin{tabular}{c} %% tabular useful for creating an array of images 
   \includegraphics [width=\textwidth]%[height=7.5cm]
   {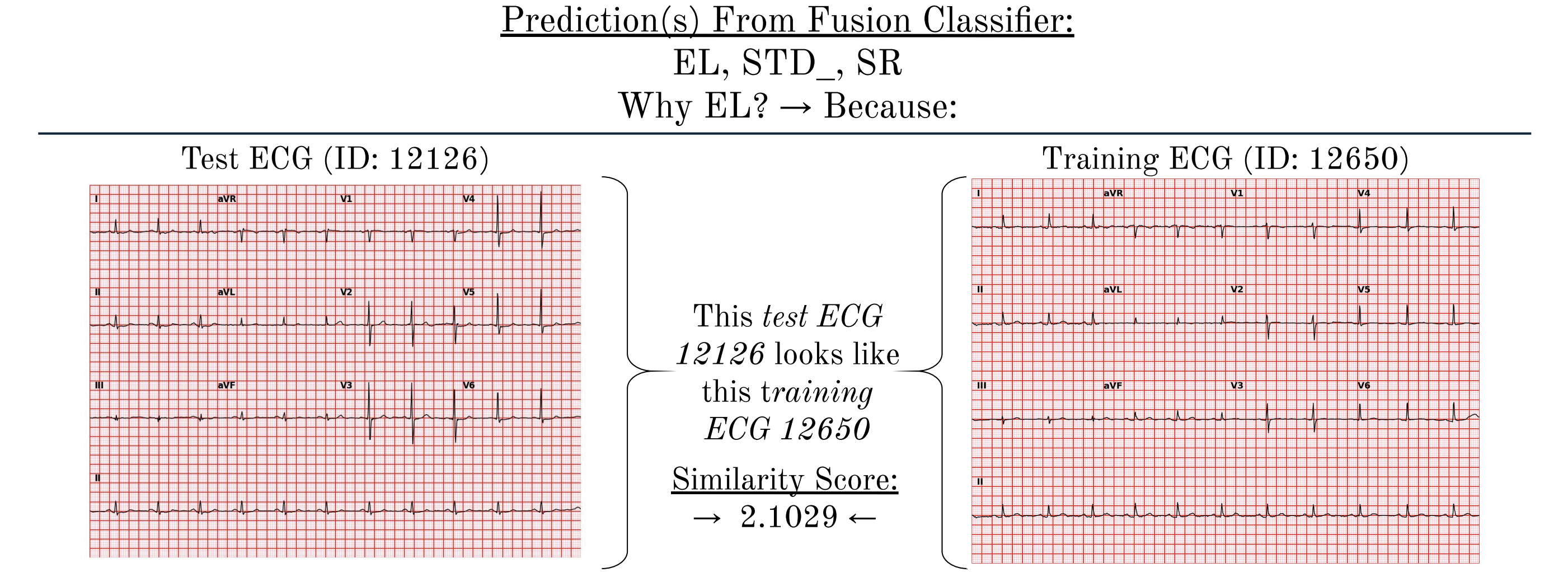}
   \end{tabular}
   \end{center}
   \caption[example] 
%>>>> use \label inside caption to get Fig. number with \autoref{}
   { \label{fig:cat4_case} Case-based explanation for an electrolyte disturbance (EL) predicted by the fusion classifier. The model predicts EL for test ECG 12126, citing strong similarity to an EL prototype that was projected onto training ECG 12650. Since this diagnosis uses 2D global prototypes, full 12-lead ECGs are shown for both the test and training examples—along with their similarity score. }
\end{figure}

\textbf{Results:} We selected one example from each branch—1D rhythm (\autoref{fig:cat1_case}), 2D local morphology (\autoref{fig:cat3_case}), and 2D global (\autoref{fig:cat4_case}). 

\textbf{Discussion:} These qualitative examples illustrate how ProtoECGNet grounds predictions in concrete, case-based reasoning for all three prototype branches. By structuring prototypes to align with rhythm, morphology, and global interpretation styles, the model produces interpretable justifications that mimic clinical reasoning. Unlike saliency maps, these explanations are faithful by design—derived from similarity to real training examples. While this analysis is qualitative, it highlights the explanatory potential of structured prototype learning for real-world decision support.

\subsection{Limitations}

One limitation of our work is that our prototype taxonomy relies on a manually defined grouping of PTB-XL labels into rhythm, morphology, and global categories. While this structure was grounded in clinical ECG interpretation heuristics, it introduces simplifications that may not fully capture the nuances of certain diagnoses. Some conditions may exhibit features that span multiple reasoning modalities (e.g., both rhythm and morphology), which are not explicitly modeled in the current architecture. Future work could explore more flexible or data-driven grouping strategies.

Second, although our prototype model is designed to support case-based explanation, we did not evaluate its impact on clinical decision-making. Our clinician review focused on the clarity and representativeness of projected prototypes, but did not test whether access to these explanations improves trust, diagnostic accuracy, or diagnostic performance. A future blinded user study is needed to quantify how interpretability affects clinical utility.

Third, we used ResNet-based CNNs as the backbone architecture across all experiments to ensure consistency in comparison. While this design choice was appropriate for isolating the effect of prototype modeling and contrastive loss, it may limit performance relative to more recent architectures. ProtoECGNet is modular by design and compatible with alternative backbones, which should be explored to assess potential gains in performance.

Finally, this study uses only the PTB-XL dataset, which contains ECGs from a single metropolitan area (Berlin, Germany). As such, model generalizability to other populations remains untested. PTB-XL was chosen for its unique diagnostic breadth—71 labels spanning the full clinical scope of ECG interpretation—which enabled development of a method tailored to large-scale, multi-label classification. Nonetheless, external validation on datasets from other settings, even if more limited in scope, will be important in future work to assess broader applicability.

% Finally, our contrastive prototype loss relies on empirical label co-occurrence statistics derived from the PTB-XL training set. These statistics may reflect dataset-specific artifacts or biases and may not generalize across institutions or populations. Methods that learn co-occurrence-aware regularization in a more adaptive or transferable manner remain an area for future work.

\section{Conclusions}
\label{sec:conclusion}

Prototype-based learning can be scaled to complex, multi-label medical time-series tasks like ECG classification without sacrificing performance. By separating diagnostic labels into rhythm, morphology, and global categories, ProtoECGNet learned representations aligned with clinically distinct reasoning processes. Each branch learned prototypes tailored to its diagnostic task, and predictions were grounded in real training ECG segments. Our proposed contrastive loss improved performance by structuring the prototype space to reflect label co-occurrence patterns observed in real-world data. Contrastive training consistently outperformed standard prototype objectives and matched black-box baselines. Clinician ratings provided an initial indication that the resulting prototypes were clear and representative of their assigned classes. While our evaluation focused on ECGs, the architectural design and contrastive prototype formulation are broadly applicable to other structured time-series tasks in medicine. 

% Future work will explore prospective clinical validation, integration with electronic health records, and expansion to other modalities.

% ACKNOWLEDGEMENTS ONLY GO IN THE CAMERA-READY, NOT THE SUBMISSION
\acks{This work was funded in part by the National Institutes of Health, specifically grant number R00NS114850 to BKB. Additionally, we would like to thank the University of Chicago Center for Research Informatics (CRI) High-Performance Computing team. The CRI is funded by the Biological Sciences Division at the University of Chicago with additional funding provided by the Institute for Translational Medicine, CTSA grant number UL1 TR000430 from the National Institutes of Health.}

%Do NOT change font size of references or modify the bibliography style
\bibliography{references,additional}

\appendix    %>>>> this command starts appendixes
\section{Code Availability}
The supplementary material and code are available at: \url{https://github.com/bbj-lab/protoecgnet}
\section{ProtoECGNet Internal Architecture}
\label{app:architecture}
\begin{figure}
\centering
\includegraphics [width=\textwidth]{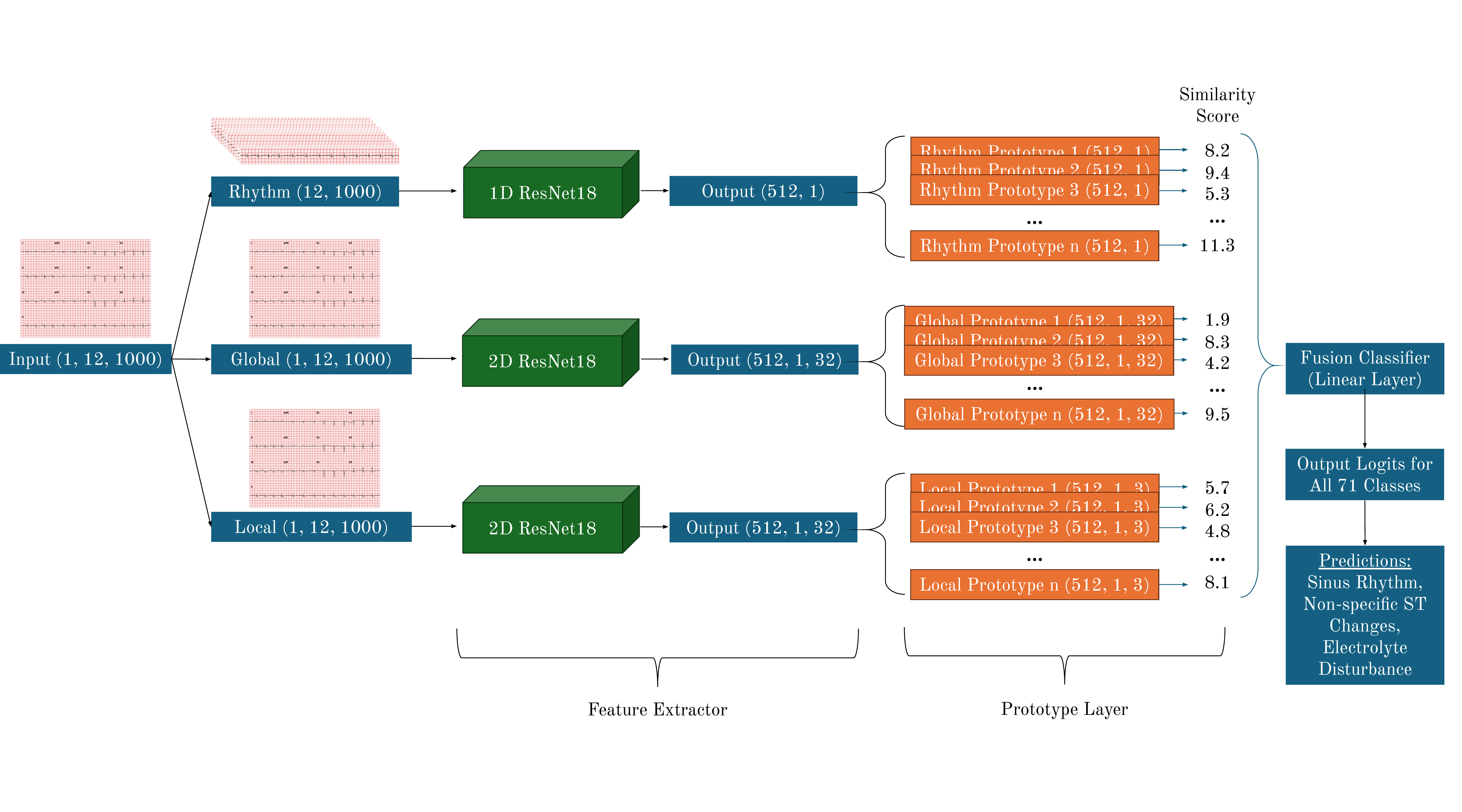}
\caption{\label{fig:app_methods} Internal ProtoECGNet architecture. Each input ECG is simultaneously passed through three branches corresponding to distinct clinical reasoning types: (1) a 1D CNN with global temporal prototypes for rhythm interpretation, (2) a 2D CNN with time-localized prototypes for morphological patterns across leads, and (3) a 2D CNN with global prototypes for diffuse signal abnormalities. Each branch is trained independently on its assigned diagnostic label subset, and outputs a similarity score for each class. A linear layer is then trained to map these similarity scores to class logits. }
\end{figure}

\autoref{fig:app_methods} contains a detailed overview of the internal architecture of each branch of ProtoECGNet.

% \section{Example of a Case-Based Explanation from the 2D Global Branch}
% \label{app:case_4_app}
% \autoref{fig:cat4_case} contains a case-based explanation for a diagnosis processed via the 2D global branch of ProtoECGNet.
% \input{figures/category4_casefig}

\section{Prototype Review Interface}
\label{app:interface}
\begin{figure}[H]
   \centering
   \includegraphics[width=0.9\linewidth]{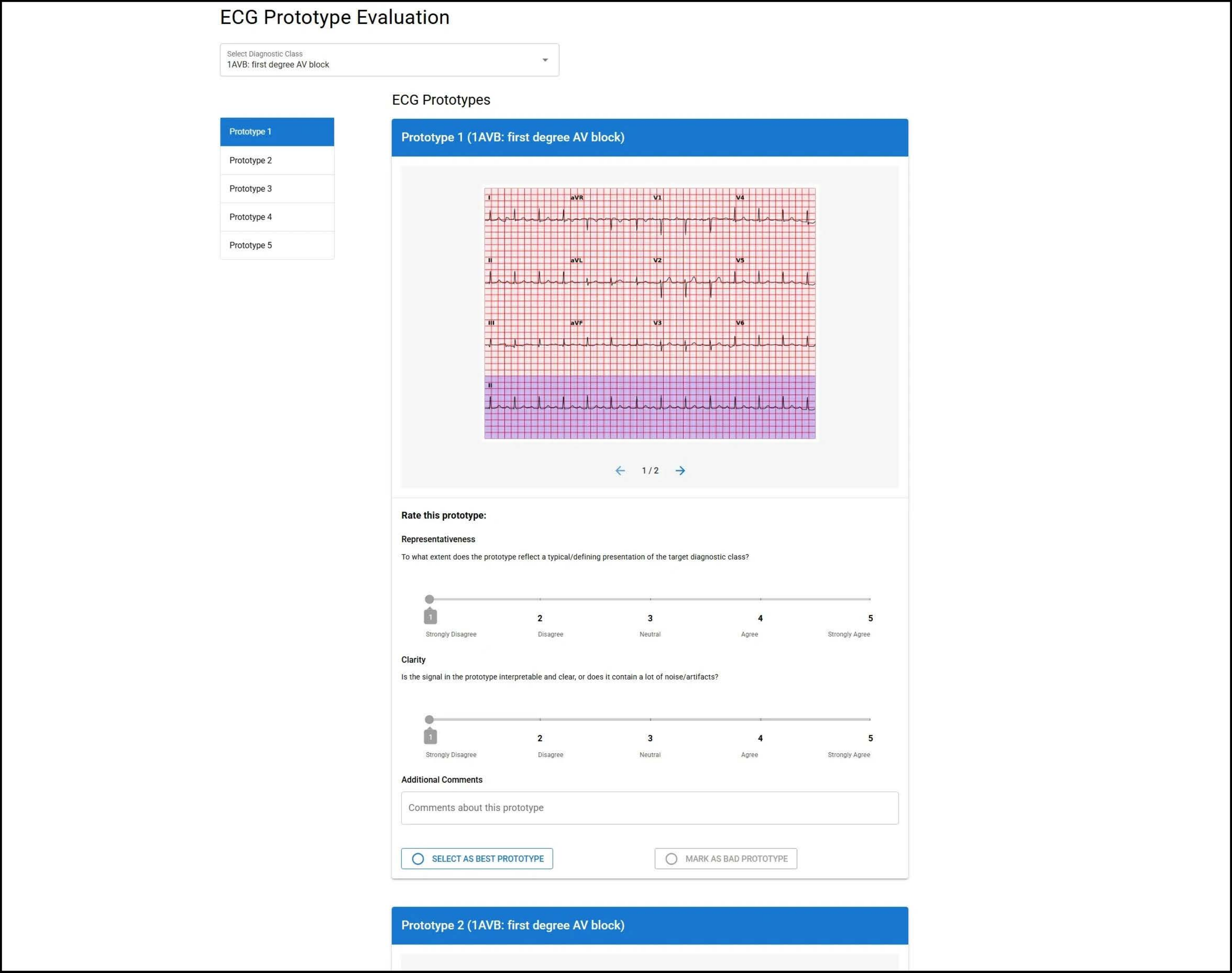}
   \caption{\label{fig:evaluation_interface} Clinician review interface for prototype evaluation.}
\end{figure}

We developed a lightweight web interface to display projected prototypes for clinician evaluation. Each prototype was rendered as a traditional 12-lead ECG with a red grid and standard calibration, labeled only by its diagnostic class. Reviewers scored each prototype independently across multiple dimensions using dropdown menus and selection boxes. A screenshot of the interface is shown in \autoref{fig:evaluation_interface}.

\section{ECG and Prototype Visualization}
\label{app:ecg_visualization}
All ECG visualizations follow the conventional clinical 12-lead layout. The top three rows display 2.5-second segments from leads I, II, III → aVR, aVL, aVF → V1–V6 in standard order, and the bottom row shows a continuous 10-second rhythm strip from lead II. This format is widely used in clinical ECG interpretation and was applied consistently in both the clinician review interface and all prototype figures to ensure familiarity for clinicians.

For each prototype, we visualize the training ECG segment that produced the latent patch onto which the prototype was projected. All branches of the model—including the 1D rhythm branch—receive the full 10-second, 12-lead ECG as input. For global prototypes (used in the 1D rhythm and 2D global branches), we display the full ECG. In the 1D rhythm branch, as the model does not explicitly model inter-lead spatial relationships, we highlight the bottom rhythm strip (lead II) to help viewers interpret the temporal pattern recognized by the model. For partial prototypes (2D morphology branch), we highlight the exact 0.94-second window (3 of 32 latent time steps) that was selected during prototype projection. To reflect the fact that the model considers all 12 leads within this window, we also include a full-lead cutout of the selected time segment, allowing assessment of local inter-lead waveform morphology. These visualizations provide clinically grounded and semantically faithful views of the prototype segments used by the model to support its predictions.

\section{Training Procedure}
\label{sec:training}

\paragraph{Weight Initialization}
We initialize the backbones in each branch by first pre-training corresponding blackbox models (i.e. 1D ResNet18, 2D ResNet18) on all 71 labels of PTB-XL. We then initialize a ProtoECGNet, freeze the backbone and classifier layers, and train for 10-20 epochs to \enquote{warm up} the prototypes as described in \citet{barnett_improving_2024}. This initialization process is optional but can improve performance.

\paragraph{Stage 1: Joint Training.}
We trained each branch independently using the loss defined in Section \ref{sec:loss}. For the final classifier in each branch, we initialized the prototype-to-class weights $W \in \mathbb{R}^{C \times P}$ such that $W_{cp} = 1$ if prototype $p$ is assigned to class $c$, and $W_{cp} = -0.5$ otherwise, like the original ProtoPNet \cite{chen_this_2019}. 

\paragraph{Stage 2: Prototype Projection.}
Each prototype $p_j$ was projected to the latent patch $z_i$ that was most similar (under the similarity metric defined in equation \ref{eq:similarity}), among training samples with label $j$:
\begin{equation}
    p_j^{\text{updated}} = \arg\max_{z_i \in \mathcal{Z}_j} \text{S}(z_i, p_j)
\end{equation}
where \( \mathcal{Z}_j \) denotes the set of all latent patches extracted from training samples that include class \( j \) as one of their labels. For \textit{partial prototypes}, \( z_i \) represents a local region of the latent space (e.g., a short time window in the morphology branch), and similarity was computed over sliding windows. For \textit{global prototypes}, \( z_i \) corresponds to the full latent representation of the input ECG (spanning all timepoints and/or leads). In both cases, the prototype was updated to exactly match the latent patch with the highest similarity among eligible samples.

\paragraph{Stage 3: Fusion Classifier Training.}
After training and projecting all prototype branches, we froze their weights and extracted the similarity scores for each ECG $i$. These per-prototype similarity scores from the 1D rhythm, 2D morphology, and 2D global branches were concatenated into a single vector:
\begin{equation}
    \mathbf{s}_i = [\mathbf{s}_i^{\text{1D}} \parallel \mathbf{s}_i^{\text{2D-p}} \parallel \mathbf{s}_i^{\text{2D-g}}] \in \mathbb{R}^{P}
\end{equation}
where $\mathbf{s}_i \in \mathbb{R}^P$ is the full similarity profile for ECG $i$ and $P$ is the total number of prototypes across all branches. We then trained a fully connected classification layer $W_{\text{fusion}} \in \mathbb{R}^{C \times P}$ on these similarity vectors to predict multi-label diagnoses:

\begin{equation}
\mathcal{L}_{\text{fusion}} = 
\frac{1}{N} \sum_{i=1}^N 
\text{BCE}(W_{\text{fusion}} \mathbf{s}_i, \mathbf{y}_i) 
+ \lambda \sum_{c=1}^C \sum_{j : \mathbf{p}_j \notin \mathcal{P}_c} 
\left| W_{\text{fusion}}^{(c,j)} \right|
\end{equation}

This stage used binary cross-entropy loss for multi-label prediction, with L1 regularization to encourage sparsity in the learned fusion weights. The sparsity constraint promoted more selective use of prototype information. Branch-specific classifiers were trained in the same way, to assess if the fusion classifier yielded performance benefits over simple macro-averaging. Additional training and tuning details are provided in Appendices \ref{app:imp_details} and \ref{app:system_req}.  

\section{Implementation Details}
\label{app:imp_details}

All models were implemented in PyTorch. We used the Adam optimizer, and explored cosine annealing, cyclic, and step-based learning rate schedulers during hyperparameter tuning. Each model was trained for up to 200 epochs, with early stopping applied if validation macro-AUROC did not improve for 10 consecutive epochs. Model checkpoints were saved based on the best validation AUROC, and the corresponding weights were used for inference and testing.

Hyperparameter tuning was conducted using Optuna with 200 trials per model variant to optimize validation macro-AUROC. Tuned parameters included learning rate (range 1e-6 to 1e-2), weight decay (range 1e-6 to 1e-2), dropout rate (range 1e-6 to 1e-2), and learning rate scheduler (ReduceLROnPlateau, CosineAnnealingLR, or CyclicLR). Both the joint training phase and the classifier stage (including the fusion classifier and branch-specific classifiers) were tuned independently using this strategy. However, for the joint phase, we also tuned number of prototypes per class (range 1-20), and for the classifier stage, we tuned the L1 penalty (range 1e-6 to 1e-2). Our final model used five 1D rhythm prototypes, eighteen 2D partial prototypes, and seven 2D global prototypes per class.

When contrastive loss was not used, we adopted the prototype loss coefficients from \citet{barnett_improving_2024}. For contrastive models, we conducted a dedicated 1000-trial Optuna sweep to tune all four loss coefficients (range 1e-6 to 1e3). This sweep fixed non-loss hyperparameters (dropout = 0.3, batch size = 32, learning rate = 0.001, scheduler = ReduceLROnPlateau, L2 weight decay = \( 1 \times 10^{-4} \)), and only optimized the four loss weights. This was run separately for the 1D rhythm, 2D partial, and 2D global prototype branches; the resulting optimal values were relatively consistent across branches, and we selected rounded values that fell between the final tuned values of all three branches (\( \lambda_{\text{clst}}=0.004, \lambda_{\text{sep}}=0.0004, \lambda_{\text{div}}=250, \lambda_{\text{cntrst}}=300 \)). These values were then held fixed in all subsequent experiments using contrastive regularization. 

\section{System Requirements}
\label{app:system_req}

All model training and inference were conducted on a high-performance computing cluster using a single NVIDIA A100 GPU with 40GB memory. Models were implemented in Python 3.10 using PyTorch 2.6.0 and PyTorch Lightning 2.5.0. Training visualization was performed with TensorBoard. Hyperparameter optimization was conducted with Optuna 4.2.1. All code was executed in a SLURM-managed Linux environment with CUDA 12 and cuDNN 9.1.

\section{Diagnostic Groupings}
\label{app:diag_groups}

Each of the 71 PTB-XL labels was assigned to one of three prototype branches based on the type of visual reasoning required for diagnosis. These groupings were defined by two board-certified physicians (one cardiologist \& one internist) and used to train the corresponding branch of ProtoECGNet. The full list of SCP codes and their assigned groupings is shown below.

\subsection*{1D Rhythm Branch (16 diagnoses)}
\begin{itemize}
    \item 1AVB: first degree AV block
    \item 2AVB: second degree AV block
    \item 3AVB: third degree AV block
    \item AFIB: atrial fibrillation
    \item AFLT: atrial flutter
    \item BIGU: bigeminal pattern (unknown origin, SV or ventricular)
    \item IVCD: nonspecific intraventricular conduction disturbance
    \item PACE: artificial pacemaker
    \item PSVT: paroxysmal supraventricular tachycardia
    \item SARRH: sinus arrhythmia
    \item SBRAD: sinus bradycardia
    \item SR: sinus rhythm
    \item STACH: sinus tachycardia
    \item SVARR: supraventricular arrhythmia
    \item SVTAC: supraventricular tachycardia
    \item TRIGU: trigeminal pattern (unknown origin, SV or ventricular)
\end{itemize}

\subsection*{2D Morphology Branch (52 diagnoses)}
\begin{itemize}
    \item ABQRS: abnormal QRS
    \item ALMI: anterolateral myocardial infarction
    \item AMI: anterior myocardial infarction
    \item ANEUR: ST-T changes from ventricular aneurysm
    \item ASMI: anteroseptal myocardial infarction
    \item CLBBB: complete left bundle branch block
    \item CRBBB: complete right bundle branch block
    \item HVOLT: high QRS voltage
    \item ILBBB: incomplete left bundle branch block
    \item ILMI: inferolateral myocardial infarction
    \item IMI: inferior myocardial infarction
    \item INJAL: injury in anterolateral leads
    \item INJAS: injury in anteroseptal leads
    \item INJIL: injury in inferolateral leads
    \item INJIN: injury in inferior leads
    \item INJLA: injury in lateral leads
    \item INVT: inverted T waves
    \item IPLMI: inferoposterolateral myocardial infarction
    \item IPMI: inferoposterior myocardial infarction
    \item IRBBB: incomplete right bundle branch block
    \item ISCAL: ischemia in anterolateral leads
    \item ISCAN: ischemia in anterior leads
    \item ISCAS: ischemia in anteroseptal leads
    \item ISCIL: ischemia in inferolateral leads
    \item ISCIN: ischemia in inferior leads
    \item ISCLA: ischemia in lateral leads
    \item ISC\_: nonspecific ischemia
    \item LAFB: left anterior fascicular block
    \item LAO/LAE: left atrial overload/enlargement
    \item LMI: lateral myocardial infarction
    \item LNGQT: long QT interval
    \item LOWT: low amplitude T waves
    \item LPFB: left posterior fascicular block
    \item LPR: prolonged PR interval
    \item LVH: left ventricular hypertrophy
    \item LVOLT: low QRS voltage
    \item NDT: nondiagnostic T abnormalities
    \item NST\_: nonspecific ST changes
    \item NT\_: nonspecific T wave changes
    \item PAC: premature atrial complex
    \item PMI: posterior myocardial infarction
    \item PRC(S): premature complexes
    \item PVC: premature ventricular complex
    \item QWAVE: Q waves present
    \item RAO/RAE: right atrial overload/enlargement
    \item RVH: right ventricular hypertrophy
    \item SEHYP: septal hypertrophy
    \item STD\_: ST depression
    \item STE\_: ST elevation
    \item TAB\_: T wave abnormality
    \item VCLVH: voltage criteria for LVH
    \item WPW: Wolff-Parkinson-White syndrome
\end{itemize}

\subsection*{2D Global Branch (3 diagnoses)}
\begin{itemize}
    \item DIG: digitalis effect
    \item EL: electrolyte disturbance or drug effect
    \item NORM: normal ECG
\end{itemize}

\section{Class-Specific Metrics}
\label{app:class_specific_aucs}
Class-specific AUROCs for the final contrastive ProtoECGNet fusion classifier are provided below. The number of samples of each class included in the PTB-XL test fold are included in parentheses after each class name. Interpret AUROCs for classes with less than 10 samples with caution. Confidence intervals were calculated by bootstrapping; they are \enquote{N/A} when only one value for a given class is present in the test fold. 
\begin{flushleft}
NDT (182): 0.918 (0.900, 0.935) \\
NST (77): 0.851 (0.813, 0.887) \\
DIG (18): 0.936 (0.902, 0.966) \\
LNGQT (11): 0.834 (0.632, 0.981) \\
NORM (963): 0.930 (0.919, 0.940) \\
IMI (267): 0.890 (0.872, 0.907) \\
ASMI (234): 0.952 (0.939, 0.963) \\
LVH (214): 0.932 (0.915, 0.948) \\
LAFB (162): 0.967 (0.957, 0.976) \\
ISC (128): 0.956 (0.940, 0.969) \\
IRBBB (112): 0.924 (0.903, 0.943) \\
1AVB (79): 0.949 (0.933, 0.963) \\
IVCD (79): 0.656 (0.591, 0.718) \\
ISCAL (66): 0.937 (0.920, 0.952) \\
CRBBB (54): 0.997 (0.996, 0.999) \\
CLBBB (54): 0.996 (0.992, 0.999) \\
ILMI (48): 0.930 (0.880, 0.969) \\
LAO/LAE (42): 0.756 (0.680, 0.825) \\
AMI (35): 0.865 (0.810, 0.915) \\
ALMI (27): 0.957 (0.924, 0.981) \\
ISCIN (22): 0.928 (0.886, 0.963) \\
INJAS (22): 0.989 (0.981, 0.995) \\
LMI (20): 0.867 (0.804, 0.923) \\
ISCIL (18): 0.950 (0.921, 0.975) \\
LPFB (18): 0.935 (0.865, 0.984) \\
ISCAS (17): 0.972 (0.954, 0.985) \\
INJAL (14): 0.974 (0.942, 0.995) \\
ISCLA (13): 0.893 (0.832, 0.949) \\
RVH (12): 0.974 (0.957, 0.988) \\
ANEUR (10): 0.952 (N/A) \\
RAO/RAE (10): 0.951 (N/A) \\
EL (9): 0.836 (0.668, 0.951) \\
WPW (8): 0.881 (N/A) \\
ILBBB (8): 0.895 (N/A) \\
IPLMI (5): 0.772 (N/A) \\
ISCAN (4): 0.931 (N/A) \\
IPMI (3): 0.826 (N/A) \\
SEHYP (2): 0.989 (N/A) \\
INJIN (2): 0.996 (N/A) \\
INJLA (2): 0.800 (N/A) \\
PMI (2): 0.995 (N/A) \\
3AVB (2): 0.984 (N/A) \\
INJIL (2): 0.881 (N/A) \\
2AVB (1): 0.980 (N/A) \\
ABQRS (322): 0.781 (0.753, 0.808) \\
PVC (114): 0.980 (0.971, 0.987) \\
STD (101): 0.853 (0.818, 0.884) \\
VCLVH (87): 0.893 (0.864, 0.917) \\
QWAVE (55): 0.831 (0.781, 0.878) \\
LOWT (44): 0.904 (0.877, 0.929) \\
NT (42): 0.917 (0.886, 0.945) \\
PAC (40): 0.862 (0.803, 0.914) \\
LPR (34): 0.959 (0.943, 0.973) \\
INVT (29): 0.952 (0.929, 0.970) \\
LVOLT (18): 0.916 (0.871, 0.957) \\
HVOLT (6): 0.965 (N/A) \\
TAB (3): 0.864 (N/A) \\
STE (3): 0.669 (N/A) \\
PRC(S) (1): 0.867 (N/A) \\
SR (1674): 0.892 (0.875, 0.908) \\
AFIB (152): 0.982 (0.967, 0.993) \\
STACH (82): 0.981 (0.968, 0.990) \\
SARRH (77): 0.895 (0.862, 0.922) \\
SBRAD (64): 0.938 (0.901, 0.966) \\
PACE (28): 0.977 (0.942, 0.997) \\
SVARR (14): 0.829 (0.754, 0.900) \\
BIGU (8): 0.979 (N/A) \\
AFLT (7): 0.968 (N/A) \\
SVTAC (3): 0.995 (N/A) \\
PSVT (2): 0.999 (N/A) \\
TRIGU (2): 0.903 (N/A) \\
\end{flushleft}

% \section*{Appendix}

% Some more details about those methods, so we can actually reproduce
% them.  After the blind review period, you could link to a repository
% for the code also.  \emph{MLHC values both rigorous evaluation as well
%   as reproduciblity.}

\end{document}